\documentclass[10pt,twocolumn,letterpaper]{article}

\usepackage{cvpr}
\usepackage{times}
\usepackage{epsfig}
\usepackage{graphicx}
\usepackage{amsmath}
\usepackage{amssymb}
\usepackage{subcaption}
\usepackage{booktabs}
\usepackage{tabularx}
\usepackage{nicefrac}

\usepackage{gensymb}

\cvprfinalcopy 

\def\cvprPaperID{9345} 

\begin{document}

\title{Self-Supervised Feature Learning via Limited Context Inpainting }
\title{Self-Supervised Feature Learning by Recognizing Image Transformations }
\title{A General Framework to Guide Feature Learning Beyond Local Pixel Statistics}
\title{A General Framework to Steer Feature Learning Beyond Local Pixel Statistics}
\title{Steering Self-Supervised Feature Learning Beyond Local Pixel Statistics}
\title{Self-Supervised Feature Learning Beyond Local Pixel Statistics}
\title{Steering Feature Learning Beyond Local Pixel Statistics}
\title{Steering Self-Supervised Feature Learning Beyond Local Pixel Statistics}


\author{Simon Jenni$^1$ \qquad Hailin Jin$^2$ \qquad Paolo Favaro$^1$\\
University of Bern$^1$ \qquad Adobe Research$^2$ \\
{\tt\small \{simon.jenni,paolo.favaro\}@inf.unibe.ch  \qquad   hljin@adobe.com }     }

\maketitle

\begin{abstract}
We introduce a novel principle for self-supervised feature learning based on the discrimination of specific transformations of an image. 
We argue that the generalization capability of learned features depends on what image neighborhood size is sufficient to discriminate different image transformations: The larger the required neighborhood size and the more global the image statistics that the feature can describe. An accurate description of global image statistics allows to better represent the shape and configuration of objects and their context, which ultimately generalizes better to new tasks such as object classification and detection.
This suggests a criterion to choose and design image transformations. 
Based on this criterion, we introduce a novel image transformation that we call limited context inpainting (LCI).
This transformation inpaints an image patch conditioned only on a small rectangular pixel boundary (the limited context). Because of the limited boundary information, the inpainter can learn to match local pixel statistics, but is unlikely to match the global statistics of the image.
We claim that the same principle can be used to justify the performance of transformations such as image rotations and warping.
Indeed, we demonstrate experimentally that learning to discriminate transformations such as LCI, image warping and rotations, yields features with state of the art generalization capabilities on several datasets such as Pascal VOC, STL-10, CelebA, and ImageNet. Remarkably, our trained features achieve a performance on Places on par with features trained through supervised learning with ImageNet labels. 
\end{abstract}

\begin{figure}[t!]
    \centering
    \includegraphics[width=0.157\linewidth]{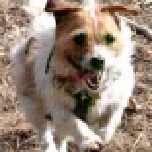}
    \includegraphics[width=0.157\linewidth]{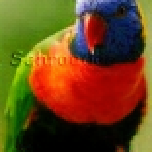}
    \includegraphics[width=0.157\linewidth]{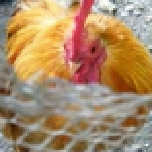}
    \includegraphics[width=0.157\linewidth]{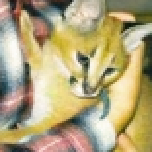}
    \includegraphics[width=0.157\linewidth]{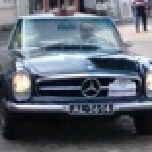}
    \includegraphics[width=0.157\linewidth]{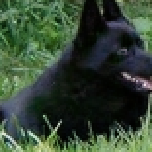}

    \includegraphics[width=0.157\linewidth]{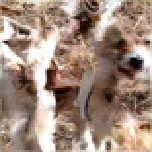}
    \includegraphics[width=0.157\linewidth]{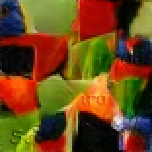}
    \includegraphics[width=0.157\linewidth]{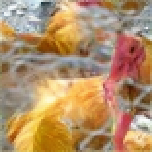}
    \includegraphics[width=0.157\linewidth]{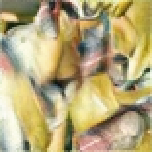}
    \includegraphics[width=0.157\linewidth]{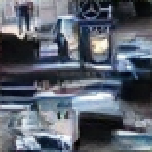}
    \includegraphics[width=0.157\linewidth]{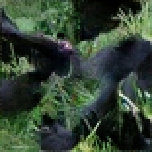}
    \caption[]{\textbf{The importance of global image statistics}. Top row: Natural images. Bottom row: Images transformed such that local statistics are preserved while global statistics are significantly altered.\protect\footnotemark An accurate image representation should be able to distinguish these two categories. A linear binary classifier trained to distinguish original versus transformed images on top of \texttt{conv5} features pre-trained on ImageNet labels yields an accuracy of 78\%. If instead we use features pre-trained with our proposed self-supervised learning task the classifier achieves an accuracy of 85\%. Notice that this transformation was not used in the training of our features and that the transformed images were built independently of either feature.}
    \label{fig:texture_vs_shape}
\end{figure}

\begin{figure}[t!]
    \centering
    \includegraphics[width=\linewidth]{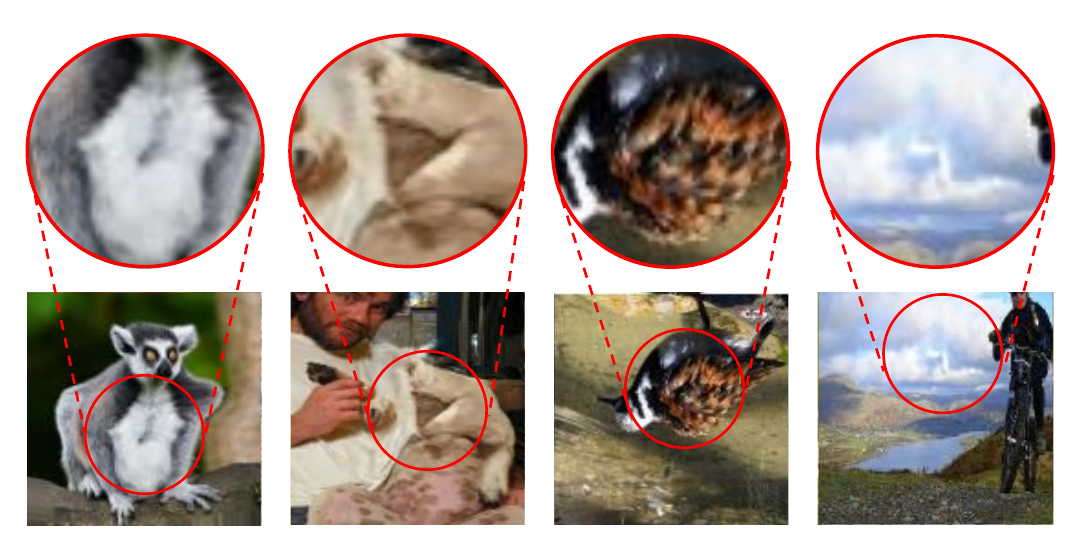}\\
    \includegraphics[width=0.22\linewidth]{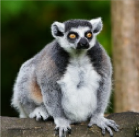}\hspace{1mm}
    \includegraphics[width=0.22\linewidth,height=0.217\linewidth]{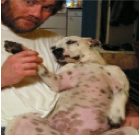}\hspace{1mm}
    \includegraphics[width=0.22\linewidth,height=0.217\linewidth]{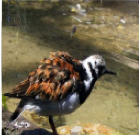}\hspace{1mm}
    \includegraphics[width=0.22\linewidth,height=0.217\linewidth]{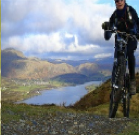}\\
    \hspace{.02\linewidth}
    (a)\hspace{.185\linewidth}
    (b)\hspace{.185\linewidth}
    (c)\hspace{.185\linewidth}
    (d)\hspace{.05\linewidth}
    \caption{\textbf{Selected image transformations}. Examples of local patches from images that were (a) warped, (b) locally inpainted, (c) rotated or (d) not transformed. The bottom row shows the original images, the middle row shows the corresponding transformed images and the top row shows a detail of the transformed image. 
    By only observing a local patch (top row), is it possible in all of the above cases to tell if and how an image has been transformed or is it instead necessary to observe the whole image (middle row), \ie, the global pixel statistics? 
    }
    \label{fig:idea}
\end{figure}

\section{Introduction}

The top-performance approaches to solve vision-based tasks, such as object classification, detection and segmentation, are currently based on supervised learning.
Unfortunately, these methods achieve a high-performance only through a large amount of labeled data, whose collection is costly and error-prone. 
Learning through labels may also encounter another fundamental limitation, depending on the training procedure and dataset: It might yield features that describe mostly local statistics, and thus have limited generalization capabilities. 
\footnotetext{The transformed images are obtained by partitioning an image into a $4\times4$ grid, by randomly permuting the tiles, and by training a network to inpaint a band of pixels across the tiles through adversarial training \cite{goodfellow2014generative}.}
An illustration of this issue is shown in Fig.~\ref{fig:texture_vs_shape}. 
On the bottom row we show images that have been transformed such that local statistics of the corresponding image on the top row are preserved, but global statistics are not. 
We find experimentally that features pre-trained with ImageNet labels  \cite{imagenet_cvpr09} have difficulties in telling real images apart from the transformed ones. This simple test shows that the classification task in ImageNet could be mostly solved by focusing on local image statistics. Such problem might not be noticed when evaluating these features on other tasks and datasets that can be solved based on similar local statistics. However, more general classification settings would certainly expose such a limitation. 
\cite{geirhos2018imagenet} also pointed out this problem and showed that  training supervised models to focus on the global statistics (which they refer to as \emph{shape}) can improve the generalization and the robustness of the learned image representation.

Thus, to address this fundamental shortcoming and to limit the need for human annotation, we propose a novel self-supervised learning (SSL) method. 
SSL methods learn features without manual labeling and thus they have the potential to better scale their training and leverage large amounts of existing unlabeled data. 
The training task in our method is to \emph{discriminate global image statistics}.
To this end, we transform images in such a way that local statistics are largely unchanged, while global statistics are clearly altered.
By doing so, we make sure that the discrimination of such transformations is not possible by working on just local patches, but instead it requires using the whole image. 
We illustrate this principle in Fig.~\ref{fig:idea}.
Incidentally, several existing SSL tasks can be seen as learning from such transformations, \eg, spotting artifacts \cite{jenni2018self}, context prediction \cite{pathak2016context}, rotation prediction \cite{gidaris2018unsupervised}, and solving jigsaw puzzles \cite{noroozi2016unsupervised}.

\begin{figure}[]
    \centering
    \includegraphics[width=\linewidth]{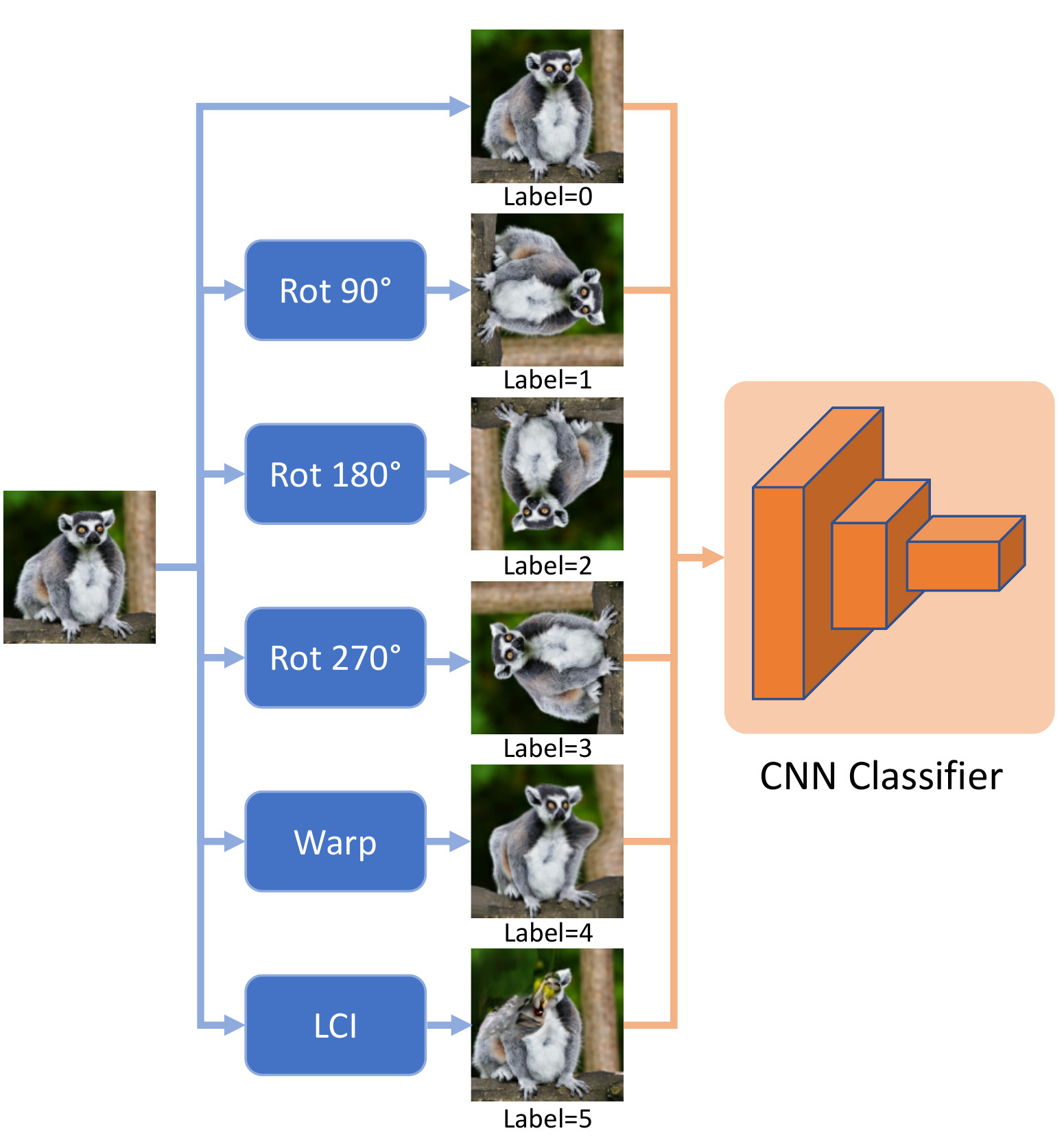}
    \caption{\textbf{Learning global statistics}. We propose to learn image representations by training a convolutional neural network to classify image transformations. The transformations are chosen such that local image statistics are preserved while global statistics are distinctly altered.}
    \label{fig:model_full}
\end{figure}

We cast our self-supervised learning approach as the task of discriminating changes in the global image statistics by classifying several image transformations (see Fig.~\ref{fig:model_full}).
As a novel image transformation we introduce \emph{limited context inpainting} (LCI).
LCI selects a random patch from a natural image, substitutes the center with noise (thus, it preserves a small outer boundary of pixels), and trains a network to inpaint a realistic center through adversarial training. While LCI can inpaint a realistic center of the patch so that it seamlessly blends with the preserved boundaries, it is unlikely to provide a meaningful match with the rest of the original image. Hence, this mismatch can only be detected by learning global statistics of the image.
Our formulation is also highly scalable and allows to easily incorporate more transformations as additional categories.
In fact, we also include the classification of image warping and image rotations (see examples of such transformations in Fig.~\ref{fig:idea}).
An illustration of the proposed training scheme is shown in Fig.~\ref{fig:model_full}.\\
\\
\noindent\textbf{Contributions.} 
Our proposed method has the following original contributions: 1) We introduce a novel self-supervised learning principle based on  image transformations that can be detected only through global observations; 2) We introduce a novel transformation according to this principle and demonstrate experimentally its impact on feature learning; 3) We formulate the method so that it can easily scale with additional transformations; 4) Our proposed method achieves state of the art performance in transfer learning on several data sets; in particular, for the first time, we show that our trained features when transferred to Places achieve a performance on par with features trained through supervised learning with ImageNet labels. Code is available at \url{https://sjenni.github.io/LCI}.


\section{Prior Work}

\noindent \textbf{Self-supervised Learning.}
Self-supervised learning is a feature learning method that avoids the use of data labels by introducing an artificial task.  Examples of tasks defined on images are to find: the  spatial configuration of parts  \cite{doersch2015unsupervised,noroozi2016unsupervised,nathan2018improvements}, the color of a grayscale image \cite{zhang2016colorful, zhang2016split, larsson2017colorproxy}, the image patch given its context \cite{pathak2016context}, the image orientation \cite{gidaris2018unsupervised}, the artifacts introduced by a corruption process \cite{jenni2018self}, the image instance up to data jittering \cite{dosovitskiy2014discriminative, wu2018unsupervised, ye2019unsupervised}, contrastive predictive coding \cite{oord2018representation,henaff2019data} or pseudo-labels obtained from a clustering process \cite{noroozi2018boosting, caron2018deep, zhuang2019local}. Self-supervised learning has also been applied to other data domains such as video \cite{wang2015unsupervised, pathakCVPR17learning, vondrick2018tracking, misra2016shuffle} and audio \cite{owens2016ambient, zhao2018sound,gao2018learning}.

Several self-supervised tasks can be seen as the prediction of some form of image transformation applied to an image. Gidaris \etal \cite{gidaris2018unsupervised} for example predict the number of 90\degree \  rotations applied to an image. Jenni and Favaro \cite{jenni2018self} predict the presence and position of artifacts introduced by a corruption process. Doersch \etal \cite{doersch2015unsupervised} predict transformations concerning image patches by predicting their relative location.  Noroozi and Favaro~\cite{noroozi2016unsupervised} extend this idea to multiple patches by solving jigsaw puzzles. Recently Zhang \etal \cite{zhang2019aet} proposed to predict the parameters of a relative projective transformation between two images using a Siamese architecture. In our work, we show that by predicting a combination of novel and previously explored image transformations we can form new and more challenging learning tasks that learn better features.  

Some works have explored the combination of different self-supervised tasks via multi-task learning \cite{ren2018cross,doersch2017multi}. Recently, Feng \etal \cite{Feng_2019_CVPR} showed that a combination of the rotation prediction task by Gidaris \etal \cite{gidaris2018unsupervised} with the instance recognition task by Wu \etal \cite{wu2018unsupervised} achieve state-of-the-art results in transfer experiments. They do so by splitting the penultimate feature vector into two parts: One to predict the transformation and a second transformation agnostic part, used to discriminate between different training images. Note that our work is orthogonal to these approaches and thus it could be integrated in such multi-task formulations and would likely lead to further improvements.

Because in our LCI transformation we build an inpainting network through adversarial training, we briefly discuss works that exploit similar techniques.\\
\noindent \textbf{Adversarial Feature Learning.} 
Generative Adversarial Networks (GANs) \cite{goodfellow2014generative} have been used for the purpose of representation learning in several works. Radford \etal \cite{radford2015unsupervised} first showed that a convolutional discriminator can learn reasonably good features.   
Donahue \etal~\cite{donahue2016adversarial,donahue2019large} learn features by training an encoder to produce the inverse mapping of the generator.
Pathak \etal~\cite{pathak2016context} use an adversarial loss to train an autoencoder for inpainting. They use the trained encoder as a feature extractor. Denton \etal \cite{denton2016semi} also perform inpainting, but instead transfer the discriminator features.
The work by Jenni and Favaro \cite{jenni2018self} has some similarity to our LCI transformation. They generate image artifacts by erasing and locally repairing features of an autoencoder. Our limited context inpainting is different from these methods in two important ways. First, we more strongly limit the context of the inpainter and put the inpainted patch back into a larger context to produce unrealistic global image statistics. Second, a separate patch discriminator allows stable adversarial training independent of the feature learning component.

\noindent \textbf{Recognizing Image Manipulations.} Many works have considered the detection of image manipulations in the context of image forensics \cite{huh2018fighting, wang2019detecting, zhou2018learning, bappy2017exploiting}. For example, Wang \etal~\cite{wang2019detecting} predict subtle face image manipulations based on local warping. Zhou \etal~\cite{zhou2018learning} detect image tampering generated using semantic masks. Transformations in these cases are usually subtle and do not change the global image statistics in a predictable way (images are manipulated to appear realistic). The aim is therefore antithetical to ours. 


\section{Learning Features by Discriminating Global Image Transformations}

Our aim is to learn image representations without human annotation by recognizing variations in global image statistics. We do so by distinguishing between natural images and images that underwent several different image transformations. Our principle is to choose image transformations that: 1) Preserve local pixel statistics (\eg, texture), but alter the global image statistics of an image and 2) Can be recognized from a single transformed example in most cases. 
In this paper we choose the following transformations: limited context inpainting, warping, rotations and the identity. These transformations will be introduced in detail in the next sections.

Formally, given a set of unlabelled training images $\{x_i\}_{i=1,\ldots,N}$ and a set of image transformations $\{T_j\}_{j=0,\ldots,K}$, we train a classifier $C$ to predict the transformation-label $j$ given a transformed example $T_j \circ x_i$. In our case we set $K=5$. We include the identity (no-transformation) case by letting $T_0 \circ x\doteq x$.
We train the network $C$ by minimizing the following self-supervised objective 
\begin{equation}
    {\cal L}_\text{SSL}(T_0,\dots,T_5) \doteq \min _{C} \frac{1}{6N} \sum_{i=1}^{N} \sum_{y=0}^{5} \ell_\text{cls} \Big(C\big(T_y \circ x_i \big), y\Big),
    \label{eq:formulation}
\end{equation}
where $\ell_\text{cls}$ is the standard cross-entropy loss for a multi-class classification problem.

\begin{figure*}[t!]
    \centering
    \includegraphics[width=0.9\linewidth]{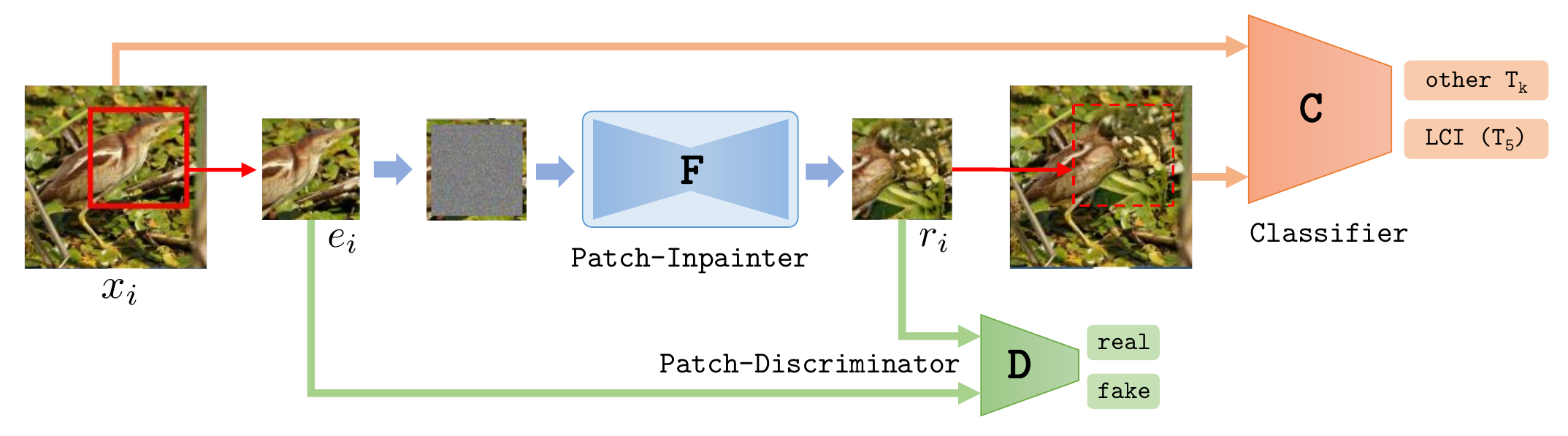}
    \caption{\textbf{Training of the Limited Context Inpainting (LCI) network}. A random patch is extracted from a training image $x$ and all but a thin border of pixels is replaced by random noise. The inpainter network $F$ fills the patch with realistic textures conditioned on the remaining border pixels. The resulting patch is replaced back into the original image, thus generating an image with natural local  statistics, but unnatural global statistics.}
    \label{fig:my_label}
\end{figure*}

\subsection{Limited Context Inpainting}

The first transformation that we propose to use in eq.~\eqref{eq:formulation} is based on the Limited Context Inpainting (LCI). The aim of LCI is to modify images only locally, \ie, at the scale of image patches. We train an inpainter network $F$ conditioned only on a thin border of pixels of the patch (see Fig.~\ref{fig:my_label}).
The inpainted patch should be realistic on its own 
and blend in at the boundary with the surrounding image, but should not meaningfully match the content of the whole image (see an example in Fig.~\ref{fig:idea}~(b)).
The inpainter $F$ is trained using adversarial training against a patch discriminator $D$ (which ensures that we match the local statistics) as well as the transformation classifier $C$. 
The patch to be inpainted is randomly selected at a uniformly sampled location $\Delta\in \Omega$, where $\Omega$ is the image domain. Then, ${\cal W}_\Delta\subset \Omega$ is a square region of pixels around $\Delta$. We define $e_i$ as the original patch of pixels at ${\cal W}_\Delta$ and $r_i$ as the corresponding inpainted patch 
\begin{align}
e_i&(p-\Delta) \doteq x_i(p),\quad \forall p\in{\cal W}_\Delta\\
r_i&\doteq F(e_i\odot (1-m) + z\odot m)
\end{align}
with $m$ a mask that is $1$ in the center of the patch and $0$ at the boundary ($2$ to $4$ pixels in our baseline), $z\sim{\cal N}(0,I)$ is a zero-mean Gaussian noise and $\odot$ denotes the Hadamard (pixel-to-pixel) product.
The LCI transformation $T_5$ is then defined as  
\begin{align}
(T_5 \circ x_i)(p) \doteq \begin{cases}
x_i(p) & \text{if } p\notin{\cal W}_\Delta\\
r_i(p-\Delta) & \text{if } p\in{\cal W}_\Delta.
\end{cases}
\end{align}
Finally, to train the inpainter $F$ we minimize the cost
\begin{align}
    {\cal L}_\text{inp} =&\frac{1}{N} \sum_{i=1}^{N} \ell_\text{GAN}(r_i,e_i)
    +\lambda_\text{border}\left| (r_i - e_i)\circ (1-m) \right|^2\nonumber\\
    &-{\cal L}_\text{SSL}(T_0,\dots,T_5),
    \label{eq:LCI}
\end{align}
where $\lambda_\text{border}=50$ is a tuning parameter to regulate the importance of autoencoding the input boundary, and $\ell_\text{GAN}(\cdot,\cdot)$ is the hinge loss for adversarial training \cite{lim2017geometric}, which also includes the maximization in the discriminator $D$.

\noindent\textbf{Remark.} 
In contrast to prior SSL methods \cite{jenni2018self,pathak2016context, denton2016semi} , here we do not take the features from the networks that we used to learn the transformation (\eg, $D$ or $F$).
Instead, here we take features from a separate classifier $C$ that has only a partial role in the training of $F$. This separation has several advantages: 1) A separate tuning of training parameters is possible, 2) GAN tricks can be applied without affecting the classifier $C$, (3) GAN training can be stable even when the classifier wins (${\cal L}_\text{SSL}$ saturates w.r.t. $F$).




\subsection{Random Warping}

In addition to the LCI, which is a local image transformation, we consider random global warping as our $T_4$ transformation. A warping is a smooth deformation of the image coordinates defined by $n$ pixel coordinates $\{(u_i, v_i)\}_{i=1,\ldots,n}$, which act as control points. We place the control points on an uniform grid of the image domain and then randomly offset each control point by sampling the shifts from a rectangular range $[-d,d]\times[-d,d]$, where $d$ is typically $\nicefrac{1}{10}$-th of the image size.
The dense flow field for warping is then computed by interpolating between the offsets at the control points using a polyharmonic spline \cite{duchon1977splines}.
Warping affects the local image statistics only minimally: In general, it is difficult to distinguish a warped patch from a patch undergoing a change in perspective. Therefore, the classifier needs to learn global image statistics to detect image warping.


\subsection{Image Rotations}

Finally, we consider as $T_1$, $T_2$, and $T_3$ image rotations of $90$\degree, $180$\degree, and $270$\degree respectively. This choice is inspired by
Gidaris \etal \cite{gidaris2018unsupervised} who proposed RotNet, a network to predict image rotations by multiples of 90\degree. This was shown to be a simple yet effective SSL pretext task. 
These transformations are predictable because the photographer bias introduces a canonical reference orientation for many natural images.
They also require global statistics as local patches of rotated images often do not indicate the orientation of the image, because similar patches can be found in the untransformed dataset. 

\begin{figure}[]
    \centering
    \begin{subfigure}[b]{0.44\linewidth}
        \centering
        \includegraphics[height=.98\linewidth]{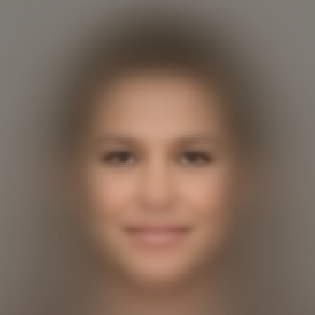}
        \caption{}
    \end{subfigure}~%
    \begin{subfigure}[b]{0.26\linewidth}
        \centering
        \includegraphics[height=0.4\linewidth,trim=2.5cm 3.5cm 4.3cm 3.5cm,clip]{figures/mean_celeba.png}\\
        \includegraphics[height=0.4\linewidth,trim=4.3cm 3.5cm 2.5cm 3.5cm,clip]{figures/mean_celeba.png}\\
        \includegraphics[height=0.4\linewidth,trim=3cm 1.3cm 3cm 5cm,clip]{figures/mean_celeba.png}\\
        \includegraphics[height=0.4\linewidth,trim=3.3cm 2.5cm 3.3cm 4.31cm,clip]{figures/mean_celeba.png}
        \caption{}
    \end{subfigure}%
    \caption{\textbf{Image statistics on CelebA}. (a) The mean image obtained from 8000 samples from CelebA. (b) Four local patches extracted from the mean image. Because these patterns appear always with the same orientation in the dataset, it is possible to distinguish rotated images by using only these local statistics.
    }
    \label{fig:mean_img}
\end{figure}

\noindent\textbf{Remark.} 
There exist, however, several settings in which the prediction of image rotations does not result in good features. Many natural images for example do not have a canonical image orientation. Thus, in these cases the prediction of image rotations is an ill-posed task. There also exist entire data domains of interest, where the image orientation is ambiguous, such as satellite and cell imaging datasets. 
Even when a clear upright image orientation exists, this method alone can lead to non-optimal feature learning. As an example, we show that the prediction of image rotations on CelebA \cite{liu2015faceattributes}, a dataset of face images, leads to significantly worse features than can be learned through the prediction of other transformations (see Table~\ref{tab:celeba_combs}). The main reason behind this limitation is that local patches can be found in the dataset always with the same orientation (see Fig.~\ref{fig:mean_img}). For instance, the classifier can easily distinguish rotated faces by simply detecting one eye or the mouth.

\subsection{Preventing Degenerate Learning}
\label{sec:degenerate}
As was observed by Doersch \etal \cite{doersch2015unsupervised}, networks trained to solve a self-supervised task might do so by using very local statistics (\eg, localization by detecting the chromatic aberration). Such solutions are called \emph{shortcuts} and are a form of degenerate learning as they yield features with poor generalization capabilities.
When introducing artificial tasks, such as the discrimination of several image transformations, it is important to make sure that the trained network cannot exploit (local) artifacts introduced by the transformations to solve the task.
For example, the classifier could learn to recognize processing artifacts of the inpainter $F$ in order to recognize LCI transformed images. 
Although adversarial training should help to prevent this behavior, we find experimentally that it is not sufficient on its own.
To further prevent such failure cases, we also train the network $F$ to autoencode image patches 
by modifying the loss ${\cal L}_\text{inp}$ in eq.~\eqref{eq:LCI} as 
$
    {\cal L}_\text{inp,AE} = {\cal L}_\text{inp} + \lambda_\text{AE}\frac{1}{N} \sum_{i=1}^{N} |F(e_i) - e_i|^2,
$
where $\lambda_\text{AE}=50$ is a tuning parameter to regulate the importance of autoencoding image patches. 
We 
create also artificial untransformed images by substituting a random patch with its autoencoded version.
In each mini-batch to the classifier we replace half of the untransformed images with these patch-autoencoded images. In this manner the classifier will not focus on the small artifacts (which could even be not visible to the naked eye) as a way to discriminate the transformations. During training we also replace half of the original images in a minibatch with these patch-autoencoded images before applying the rotation. 


\section{On the Choice of Transformations}

Our goal is to learn features by discriminating images undergoing different transformations. We pointed out that this approach should use transformations that can be distinguished only by observing large regions of pixels, and is scalable, \ie, it can be further refined by including more transformations.
In this section, we would like to make these two aspects clearer.\\
\noindent\textbf{Determining suitable transformations.}
We find that the choice of what transformations to use depends on the data distribution. An example of such dependency in the case of RotNet on CelebA is shown in Fig.~\ref{fig:mean_img}.
Intuitively, \emph{an ideal transformation is such that any transformed local patch should be found in the original dataset, but any transformed global patch should not be found in the dataset}. 
This is also the key idea behind the design of LCI.
\\
\noindent\textbf{Introducing additional transformations.} 
As we will show in the Experiments section, adding more transformations (as specified above) can improve the performance. An important aspect is that the classifier must be able to distinguish the different transformations. Otherwise, its task is ambiguous and can lead to degenerate learning. Put in simple terms, \emph{a transformed global patch should be different from any other global patch (including itself) transformed with a different transformation}. We verify that our chosen transformations satisfy this principle, as LCI and image warping cannot produce rotated images and warping is a global deformation, while LCI is a local one.


\section{Experiments}

We perform an extensive experimental evaluation of our formulation on several established unsupervised feature learning benchmarks. For a fair comparison with prior work we implement the transformation classifier $C$ with a standard AlexNet architecture \cite{krizhevsky2012imagenet}. Following prior work, we remove the local response normalization layers and add batch normalization \cite{ioffe2015batch} to all layers except for the final one. No other modifications to the original architecture were made (we preserve the two-stream architecture). 
For experiments on lower resolution images we remove the max-pooling layer after \texttt{conv5} and use SAME padding throughout the network. The standard data-augmentation strategies (random cropping and horizontal flipping) were used. 
Self-supervised pre-training of the classifier was performed using the AdamW optimizer \cite{loshchilov2018decoupled} with parameters $\beta_1=0.5$, $\beta_2=0.99$ and a weight decay of $10^{-4}$. We decayed the learning rate from $3\cdot10^{-4}$ to $3\cdot10^{-7}$ over the course of training using cosine annealing \cite{loshchilov2016sgdr}. 
The training of the inpainter network $F$ and patch-discriminator $D$ was done using the Adam optimizer \cite{kingma2014adam} with a fixed learning rate of $2\cdot10^{-4}$ and $\beta_1 = 0.5$. The size of the patch boundary is set to 2 pixels in experiments on STL-10 and CelebA. On ImageNet we use a 4 pixel boundary.
Details for the network architectures and additional results are provided in the supplementary material. 

\begin{table}[t]
\centering
\caption{Ablation experiments for different design choices of Limited Context Inpainting (LCI) on STL-10~\cite{coates2011analysis}. We pre-train an AlexNet to predict if an image has been transformed with LCI or not and transfer the frozen \texttt{conv5}  features for linear classification.}\label{tab:stl_lci}
\begin{tabular}{@{}l@{\hspace{2.5em}}c@{}}
\toprule
 \textbf{Ablation}               			& \textbf{Accuracy} \\ \midrule
(a)  $ 32 \times 32$ patches	 					&  61.2\%  \\ 
(b)  $40 \times 40$ patches	 					&  70.6\%  \\ 
(c)  $56 \times 56$ patches	 					&  75.1\%  \\ \midrule
(d)  Pre-trained and frozen $F$	 					&  63.7\%  \\
(e)  No adversarial loss w.r.t. $C$	 					&  68.0\%  \\
(f)  No patch autoencoding 					&  69.5\%  \\
\midrule
\phantom{(b)}  Baseline ($48 \times 48$ patches	)			&  76.2\%	\\
\bottomrule
\end{tabular}
\end{table}

\subsection{Ablation Experiments}

\noindent \textbf{Limited Context Inpainting.} We perform ablation experiments on STL-10~\cite{coates2011analysis} to validate several design choices for the joint inpainter and classifier training. We also illustrate the effect of the patch-size on the performance of the learned features. 
We pre-train the transformation classifier for 200 epochs on $64\times64$ crops of the unlabelled training set. The mini-batch size was set to 64. We then transfer the frozen  \texttt{conv5} features by training a linear classifier for 500 epochs on randomly cropped $96\times96$ images of the small labelled training set. Only LCI was used as transformation in these experiments. The results of the following ablations are reported in Table \ref{tab:stl_lci}:
\begin{description}
	\item [(a)-(c) Varying Patch-Size:] We vary the size of the inpainted patches. We observe that small patches lead to a significant drop in feature performance. Smaller patches are easy to inpaint and the results often do not alter the global image statistics;
	\item [(d)-(f) Preventing Shortcuts:] Following sec.~\ref{sec:degenerate}, we show how adversarial training of $F$ is necessary to achieve a good performance by removing the feedback of both $D$ and $C$ in (d) and only $C$ in (e). We also demonstrate the importance of adding autoencoded patches to the non-transformed images in (f);
\end{description}

\begin{table}[t]
\caption{We report the test set accuracy of linear classifiers trained on frozen features for models trained to predict different combinations of image transformations on STL-10.\vspace{-3mm}}\label{tab:stl10_combs}
\begin{center}
\resizebox{\linewidth}{!}{
\begin{tabular}{@{}l@{\hspace{1em}}c@{\hspace{1em}} c@{\hspace{1em}} c@{\hspace{1em}} c@{\hspace{1em}} c@{}}
\toprule
\textbf{Initialization} & \texttt{conv1} & \texttt{conv2} &  \texttt{conv3} &  \texttt{conv4} &  \texttt{conv5} \\ \midrule
 Random   &  48.4\% & 53.3\%  & 51.1\% & 48.7\%  & 47.9\% \\  \midrule
Warp   &  57.2\% & 64.2\%  & 62.8\% & 58.8\%  & 55.3\%  \\ 
LCI   &  58.8\% & 67.2\%  & 67.4\% & 68.1\%  & 68.0\%  \\ 
Rot   &  58.2\% & 67.3\%  & 69.3\% & 69.9\%  & 70.1\%  \\  \midrule
Warp + LCI   & \textbf{59.3\%} & 68.1\%  & 69.5\% & 68.5\%  & 67.2\%  \\
Rot + Warp   &  57.4\% & \underline{69.2\%}  & 70.7\% & 70.5\%  & 70.6\%  \\
Rot + LCI   &  58.5\% & \underline{69.2\%}  & \underline{71.3\%} & \underline{72.8\%}  & \underline{72.3\%}  \\  \midrule
Rot + Warp + LCI   &  \underline{59.2\%} & \textbf{69.7\%}  & \textbf{71.9\%} & \textbf{73.1\%}  & \textbf{73.7\%}  \\ 
\bottomrule
\end{tabular}}
\end{center}
\end{table}

\begin{table}[t]
\caption{We report the average precision of linear classifiers trained to predict facial attributes on frozen features of models trained to predict different combinations of image transformations on CelebA.\vspace{-3mm}}\label{tab:celeba_combs}
\begin{center}
\resizebox{\linewidth}{!}{%
\begin{tabular}{@{}l@{\hspace{1em}}c@{\hspace{1em}} c@{\hspace{1em}} c@{\hspace{1em}} c@{\hspace{1em}} c@{}}
\toprule
\textbf{Initialization} & \texttt{conv1} & \texttt{conv2} &  \texttt{conv3} &  \texttt{conv4} &  \texttt{conv5} \\ \midrule
 Random    &  68.9\% & 70.1\%  & 66.7\% & 65.3\%  & 63.2\%  \\  \midrule
Warp   &  71.7\% & 73.4\% & 71.2\% & 68.8\%  & 64.3\%  \\ 
LCI   &  71.3\% & 73.0\% & 72.0\% & 71.1\%  & 68.0\%  \\ 
Rot   &  70.3\% & 70.9\% & 67.8\% & 65.6\%  & 62.1\%  \\ \midrule
Warp + LCI   &  \textbf{72.0\%} & \underline{73.9\%}  & \underline{73.3\%} & \underline{72.1\%}  & \underline{69.0\%}  \\
Rot + Warp   &  71.6\% & 73.6\% & 72.0\% & 70.1\%  & 66.4\%  \\ 
Rot + LCI   &  71.3\% & 72.7\%  & 71.9\% & 70.8\%  & 66.7\%  \\ \midrule
Rot + Warp + LCI   &  \underline{71.8\%} & \textbf{74.0\%}  & \textbf{73.5\%} & \textbf{72.5\%}  & \textbf{69.2\%}  \\ 
\bottomrule
\end{tabular}}
\end{center}
\end{table}

\noindent \textbf{Combination of Image Transformations.} We perform additional ablation experiments on STL-10 and CelebA~\cite{liu2015faceattributes} where $C$ is trained to predict different combinations of image transformations. These experiments illustrate how our formulation can scale with the number of considered image transformations and how the effectiveness of transformations can depend on the data domain.   

We pre-train the AlexNet to predict image transformations for 200 epochs on $64\times64$ crops on STL-10 and for 100 epochs on $96\times96$ crops on CelebA using the standard data augmentations. For transfer we train linear classifiers on top of the frozen convolutional features (without resizing of the feature-maps) to predict the 10 object categories in the case of STL-10 and to predict the 40 face attributes in the case of CelebA. Transfer learning is performed for 700 epochs on $64\times64$ crops in the case of STL-10 and for 100 epochs on $96\times96$ crops in the case of CelebA. We report results for STL-10 in Table \ref{tab:stl10_combs} and for CelebA in Table \ref{tab:celeba_combs}.

We can observe that the discrimination of a larger number of image transformations generally leads to better feature performance on both datasets. When considering each of the transformations in isolation we see that not all of them generalize equally well to different data domains. Rotation prediction especially performs significantly worse on CelebA than on STL-10. The performance of LCI on the other hand is good on both datasets. 

\begin{table}[t]
\caption{Transfer learning results for classification, detection and segmentation on PASCAL compared to state-of-the-art feature learning methods (* use a bigger AlexNet).\vspace{-3mm}}\label{tab:voc_layers}
\begin{center}
\resizebox{\linewidth}{!}{%
\begin{tabular}{@{}l@{\hspace{.1em}}c@{\hspace{.2em}}c@{\hspace{.5em}}c@{\hspace{.5em}}c@{}}
\toprule
  &   & \textbf{Classification}  & \textbf{Detection}  & \textbf{Segmentation} \\
\textbf{Model} & \textbf{[Ref]}  & \textbf{(mAP)}  & \textbf{(mAP)}  & \textbf{(mIoU)} \\ \midrule
Krizhevsky \etal  \cite{krizhevsky2012imagenet} & \cite{zhang2016colorful} & 79.9\% & 59.1\%  &  48.0\%     \\ 
Random & \cite{pathak2016context} & 53.3\% & 43.4\% & 19.8\%     \\  \hline
Agrawal \etal \cite{agrawal2015learning} &	\cite{donahue2016adversarial} & 54.2\% & 43.9\% & -  \\
Bojanowski \etal \cite{bojanowski2017unsupervised} &	 \cite{bojanowski2017unsupervised} & 65.3\% & 49.4\% & -  \\
Donahue \etal \cite{donahue2016adversarial} &	\cite{donahue2016adversarial} & 60.1\% & 46.9\% & 35.2\%        \\
Feng \etal \cite{Feng_2019_CVPR} & \cite{Feng_2019_CVPR} & \underline{74.3\%}    &   \textbf{57.5\%}   & \textbf{45.3\%}           \\
Gidaris \etal \cite{gidaris2018unsupervised} & \cite{gidaris2018unsupervised} & 73.0\%    &   54.4\%   &  39.1\%           \\
Jayaraman \& Grauman \cite{jayaraman2015learning} & \cite{jayaraman2015learning} & - & 41.7\% & -          \\
Jenni \& Favaro \cite{jenni2018self} & \cite{jenni2018self} & 69.8\%    &  52.5\%   & 38.1\%           \\
Kr\"ahenb\"uhl \etal \cite{krahenbuhl2015data} & \cite{krahenbuhl2015data} & 56.6\% & 45.6\% & 32.6\%          \\
Larsson \etal \cite{larsson2017colorproxy} & \cite{larsson2017colorproxy} & 65.9\% & - & 38.0\%          \\
Noroozi \& Favaro \cite{noroozi2016unsupervised} & \cite{noroozi2016unsupervised} & 67.6\% & 53.2\% & 37.6\%   \\
Noroozi \etal \cite{noroozi2017representation} & \cite{noroozi2017representation} & 67.7\% & 51.4\% & 36.6\%   \\
Noroozi \etal \cite{noroozi2018boosting} & \cite{noroozi2018boosting} & 72.5\% & 56.5\% & 42.6\%   \\
Mahendran \etal \cite{mahendran2018cross} & \cite{mahendran2018cross} & 64.4\% & 50.3\% & 41.4\%   \\
Mundhenk \etal \cite{nathan2018improvements}  & \cite{nathan2018improvements} & 69.6\% & 55.8\% & 41.4\%   \\
Owens \etal \cite{owens2016ambient} &	\cite{owens2016ambient} & 61.3\% & 44.0\% & -          \\
Pathak \etal \cite{pathak2016context} &	\cite{pathak2016context}	 & 56.5\% & 44.5\% & 29.7\%          \\
Pathak \etal \cite{pathakCVPR17learning} &	\cite{pathakCVPR17learning}	 & 61.0\% & 52.2\% & -          \\
Wang \& Gupta \cite{wang2015unsupervised} &	 \cite{krahenbuhl2015data}	 & 63.1\% & 47.4\% & -    \\
Zhan \etal \cite{zhan2019self}  &	\cite{zhan2019self}	 & - & - & \underline{44.5\%}       \\
Zhang \etal \cite{zhang2016colorful} &	\cite{zhang2016colorful}	 & 65.9\% & 46.9\% & 35.6\%       \\
Zhang \etal \cite{zhang2016split} &	\cite{zhang2016split} & 67.1\%  & 46.7\% & 36.0\%       \\ \midrule
Doersch \etal \cite{doersch2015unsupervised}* & \cite{donahue2016adversarial}	 & 65.3\% & 51.1\% & -      \\
Caron \etal \cite{caron2018deep}* & \cite{caron2018deep}	 & 73.7\% & 55.4\% & 45.1      \\\midrule
Ours   &    -    &   \textbf{74.5\%}    &   \underline{56.8\%}   &      44.4    \\ \bottomrule
\end{tabular}}
\end{center}
\end{table}

\subsection{Unsupervised Feature Learning Benchmarks}

We compare our proposed model to state-of-the-art methods on the established feature learning benchmarks. We pre-train the transformation classifier for 200 epochs on the ImageNet training set. Images were randomly cropped to $128\times128$ and the last max-pooling layer was removed during pre-training to preserve the size of the feature map before the fully-connected layers. We used a batch-size of 96 and trained on 4 GPUs.  

\begin{table}[t]
\caption{Validation set accuracy on ImageNet with linear classifiers trained on frozen convolutional layers. $^\dagger$ indicates multi-crop evaluation and * use a bigger AlexNet.\vspace{-3mm}}\label{tab:imnet_layers}
\begin{center}
\resizebox{\linewidth}{!}{%
\begin{tabular}{@{}l@{\hspace{1em}}c@{\hspace{1em}} c@{\hspace{1em}} c@{\hspace{1em}} c@{\hspace{1em}} c@{}}
\toprule
\textbf{Model\textbackslash Layer} & \texttt{conv1} & \texttt{conv2} &  \texttt{conv3} &  \texttt{conv4} &  \texttt{conv5} \\ \midrule
ImageNet Labels   & 19.3\% & 36.3\% & 44.2\% & 48.3\%  & 50.5\%  \\
Random  & 11.6\% & 17.1\% & 16.9\% & 16.3\%  & 14.1\%  \\ \midrule
Donahue \etal \cite{donahue2016adversarial} & 17.7\% & 24.5\% & 31.0\% & 29.9\%  & 28.0\%  \\
Feng \etal \cite{Feng_2019_CVPR} & 19.3\% & \underline{33.3\%} & \textbf{40.8\%} & \underline{41.8\%}  & \textbf{44.3\%}  \\
Gidaris \etal \cite{gidaris2018unsupervised}  & 18.8\% & 31.7\%  & 38.7\% & 38.2\%  & 36.5\%  \\
Huang \etal \cite{huang2019unsupervised}  & 15.6\% & 27.0\%  & 35.9\% & 39.7\%  & 37.9\%  \\
Jenni \& Favaro \cite{jenni2018self}  & \underline{19.5\%} & 33.3\%  & 37.9\% & 38.9\%  & 34.9\%  \\
Noroozi \& Favaro \cite{noroozi2016unsupervised} & 18.2\% & 28.8\% & 34.0\% & 33.9\%  & 27.1\%  \\
Noroozi \etal \cite{noroozi2017representation} & 18.0\% & 30.6\% & 34.3\% & 32.5\%  & 25.7\%  \\  
Noroozi \etal \cite{noroozi2018boosting} & 19.2\% & 32.0\% & 37.3\% & 37.1\%  & 34.6\%  \\  
Tian \etal \cite{tian2019contrastive} & 18.4\% & 33.5\% & 38.1\% & 40.4\%  & \underline{42.6\%}  \\
Wu \etal \cite{wu2018unsupervised} & 16.8\% & 26.5\% & 31.8\% & 34.1\%  & 35.6\%  \\
Zhang \etal \cite{zhang2016colorful} & 13.1\% & 24.8\% & 31.0\% & 32.6\%  & 31.8\%  \\
Zhang \etal \cite{zhang2016split} & 17.7\% & 29.3\% & 35.4\% & 35.2\%  & 32.8\%  \\ 
Zhang \etal \cite{zhang2019aet} & 19.2\% & 32.8\% & \underline{40.6\%} & 39.7\%  & 37.7\%  \\ \midrule
Doersch \etal \cite{doersch2015unsupervised}* & 16.2\% & 23.3\% & 30.2\% & 31.7\%  & 29.6\%  \\
Caron \etal \cite{caron2018deep}* & 12.9\% & 29.2\% & 38.2\% & 39.8\%  & 36.1\%  \\
Zhuang \etal \cite{zhuang2019local}*$^\dagger$  & 18.7\% & 32.7\% & 38.1\% & 42.3\%  & 42.4\%  \\

\midrule
Ours   &  \textbf{20.8\%} & \textbf{34.5\%}  & 40.2\% & \textbf{43.1\%}  & 41.4\%  \\ 
 Ours$^\dagger$     & 22.0\% & 36.4\%  & 42.4\% & 45.4\%  & 44.4\%  \\ \bottomrule

\end{tabular}}
\end{center}
\end{table}

\begin{table}[t]
\caption{Validation set accuracy on Places with linear classifiers trained on frozen convolutional layers. $^\dagger$ indicates multi-crop evaluation and * the use of a bigger AlexNet.\vspace{-3mm}}\label{tab:places_layers}
\begin{center}
\resizebox{\linewidth}{!}{%
\begin{tabular}{@{}l@{\hspace{1em}}c@{\hspace{1em}} c@{\hspace{1em}} c@{\hspace{1em}} c@{\hspace{1em}} c@{}}
\toprule
\textbf{Model\textbackslash Layer} & \texttt{conv1} & \texttt{conv2} &  \texttt{conv3} &  \texttt{conv4} &  \texttt{conv5} \\ \midrule
Places Labels  & 22.1\% & 35.1\% & 40.2\% & 43.3\%  & 44.6\%  \\
ImageNet Labels  & 22.7\% & 34.8\% & 38.4\% & 39.4\%  & 38.7\%  \\
Random  & 15.7\% & 20.3\% & 19.8\% & 19.1\%  & 17.5\%  \\ \midrule
Donahue \etal \cite{donahue2016adversarial} & 22.0\% & 28.7\% & 31.8\% & 31.3\%  & 29.7\%  \\
Feng \etal \cite{Feng_2019_CVPR} & 22.9\% & 32.4\% & 36.6\% & \underline{37.3\%}  & \textbf{38.6\%}  \\
Gidaris \etal \cite{gidaris2018unsupervised}  & 21.5\% & 31.0\%  & 35.1\% & 34.6\%  & 33.7\%  \\
Jenni \& Favaro \cite{jenni2018self}  & \underline{23.3\%} & \textbf{34.3\%}  & 36.9\% & \underline{37.3\%}  & 34.4\%  \\
Noroozi \& Favaro \cite{noroozi2016unsupervised} & 23.0\% & 31.9\% & 35.0\% & 34.2\%  & 29.3\%  \\
Noroozi \etal \cite{noroozi2017representation} & \underline{23.3\%} & 33.9\% & 36.3\% & 34.7\%  & 29.6\%  \\  
Noroozi \etal \cite{noroozi2018boosting} & 22.9\% & \underline{34.2\%} & \underline{37.5\%} & 37.1\%  & 34.4\%  \\  
Owens \etal \cite{owens2016ambient} & 19.9\% & 29.3\% & 32.1\% & 28.8\%  & 29.8\%  \\
Pathak \etal \cite{pathak2016context} & 18.2\% & 23.2\% & 23.4\% & 21.9\%  & 18.4\%  \\
Wu \etal \cite{wu2018unsupervised} & 18.8\% & 24.3\% & 31.9\% & 34.5\%  & 33.6\%  \\
Zhang \etal \cite{zhang2016colorful} & 16.0\% & 25.7\% & 29.6\% & 30.3\%  & 29.7\%  \\
Zhang \etal \cite{zhang2016split} & 21.3\% & 30.7\% & 34.0\% & 34.1\%  & 32.5\%  \\ 
Zhang \etal \cite{zhang2019aet} & 22.1\% & 32.9\% & 37.1\% & 36.2\%  & 34.7\%  \\ \midrule
Doersch \etal \cite{doersch2015unsupervised}* & 19.7\% & 26.7\% & 31.9\% & 32.7\%  & 30.9\%  \\
Caron \etal \cite{caron2018deep}* & 18.6\% & 30.8\% & 37.0\% & 37.5\%  & 33.1\%  \\
Zhuang \etal \cite{zhuang2019local}*$^\dagger$  & 18.7\% & 32.7\% & 38.2\% & 40.3\%  & 39.5\%  \\

\midrule
Ours   &  \textbf{24.1\%} & 33.3\%  & \textbf{37.9\%} & \textbf{39.5\%}  & \underline{37.7\%}  \\ 
Ours$^\dagger$      & 25.0\% & 34.8\%  & 39.7\% & 41.1\%  & 39.4\%  \\ \bottomrule

\end{tabular}}
\end{center}
\end{table}

\noindent \textbf{Pascal VOC.} We finetune our transformation classifier features for multi-label classification, object detection and semantic segmentation on the Pascal VOC dataset. We follow the established experimental setup and use the framework provided by Kr\"ahenb\"uhl \etal~\cite{krahenbuhl2015data} for multilabel classification, the Fast-RCNN \cite{girshickICCV15fastrcnn} framework for detection and the FCN  \cite{long2015fully} framework for semantic segmentation.
We absorb the batch-normalization parameters into the parameters of the associated layers in the AlexNet and apply the data-dependent rescaling by Kr\"ahenb\"uhl \etal~\cite{krahenbuhl2015data}, as is common practice.
The results of these transfer learning experiments are reported in Table \ref{tab:voc_layers}. We achieve state-of-the-art performance in classification and competitive results for detection and segmentation. 

\noindent \textbf{Linear Classifier Experiments on ImageNet and Places.}  
To measure the quality of our self-supervised learning task we use the transformation classifier as a fixed feature extractor and train linear classifiers on top of each  convolutional layer. These experiments are performed both on ImageNet (the dataset used for pre-training) and Places \cite{NIPS2014_5349} (to measure how well the features generalize to new data). We follow the same setup as the state-of-the-art methods and report the accuracy achieved on a single crop. Results for ImageNet are shown in Table \ref{tab:imnet_layers} and for Places in Table \ref{tab:places_layers}. Our learned features achieve state-of-the-art performance for \texttt{conv1},  \texttt{conv2} and \texttt{conv4} on ImageNet. On Places we achieve the best results on \texttt{conv1},  \texttt{conv3} and  \texttt{conv4}. Our results on \texttt{conv4} in particular are the best overall and even slightly surpass the performance of an AlexNet trained on ImageNet using supervision.

\begin{figure}[]
    \centering
    \includegraphics[width=0.95\linewidth]{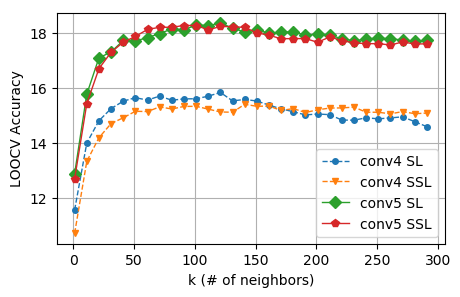}
    \caption{We report leave-one-out cross validation (LOOCV) accuracy for $k$-nearest neighbor classifiers on the Places validation set. We compare the performance of our self-supervised transformation classifier against features of a supervised network for different values of $k$. Both networks were pre-trained on ImageNet. }
    \label{fig:NN_places}
\end{figure}

\begin{figure}[]
    \centering
    \includegraphics[height=1cm,trim={0px 0px 576px 0px},clip]{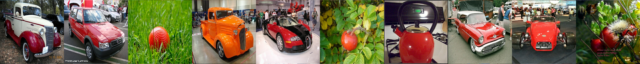}
    \includegraphics[height=1cm,trim={64px 0px 128px 0px},clip]{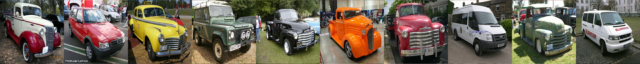}
    \hspace*{1cm} \includegraphics[height=1cm,trim={64px 0px 128px 0px},clip]{figures/nn_sl_1.png}
    \includegraphics[height=1cm,trim={0px 0px 576px 0px},clip]{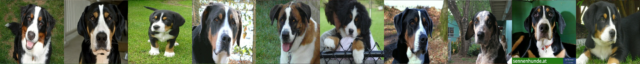}
    \includegraphics[height=1cm,trim={64px 0px 128px 0px},clip]{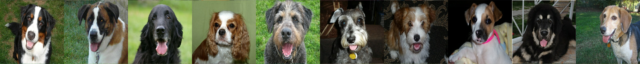}
    \hspace*{1cm} \includegraphics[height=1cm,trim={64px 0px 128px 0px},clip]{figures/nn_sl_2.png}
    \includegraphics[height=1cm,trim={0px 0px 576px 0px},clip]{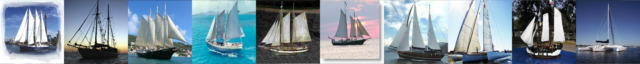}
    \includegraphics[height=1cm,trim={64px 0px 128px 0px},clip]{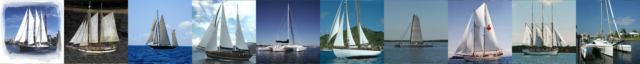}
    \hspace*{1cm} \includegraphics[height=1cm,trim={64px 0px 128px 0px},clip]{figures/nn_sl_4.png}
    \includegraphics[height=1cm,trim={0px 0px 576px 0px},clip]{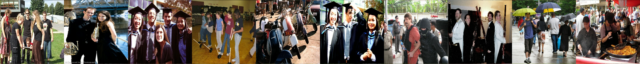}
    \includegraphics[height=1cm,trim={64px 0px 128px 0px},clip]{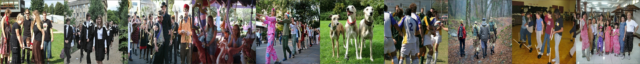}
    \hspace*{1cm} \includegraphics[height=1cm,trim={64px 0px 128px 0px},clip]{figures/nn_sl_5.png}
    \includegraphics[height=1cm,trim={0px 0px 576px 0px},clip]{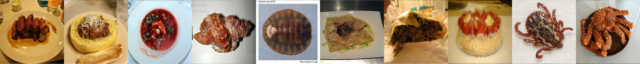}
    \includegraphics[height=1cm,trim={64px 0px 128px 0px},clip]{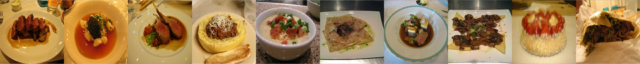}
    \hspace*{1cm} \includegraphics[height=1cm,trim={64px 0px 128px 0px},clip]{figures/nn_sl_8.png}
    \caption{Comparison of nearest neighbor retrieval. The left-most column shows the query image. Odd rows: Retrievals with our features. Even rows: Retrievals with features learned using ImageNet labels. Nearest neighbors were computed on the validation set of ImageNet with \texttt{conv5} features using cosine similarity.}
    \label{fig:NN}
\end{figure}

\noindent \textbf{Nearest Neighbor Evaluation.}
Features learned in deep CNNs through supervised learning tend to distribute so that their Euclidean distance relates closely to the semantic \emph{visual similarity} of the images they correspond to. We want to see if also our SSL features enjoy the same property.
Thus, we compute the nearest neighbors of our SSL and of SL features in \texttt{conv5} features space on the validation set of ImageNet. Results are shown in Fig.~\ref{fig:NN}.  
We also show a quantitative comparison of $k$-nearest neighbor classification on the Places validation set in Figure~\ref{fig:NN_places}.
We report the leave-one-out cross validation (LOOCV) accuracy for different values of $k$.
This can be done efficiently by computing $(k+1)$-nearest neighbors using the complete dataset and by excluding the closest neighbor for each query. 
The concatenation of features from five $128\times128$ crops (extracted at the resolution the networks were trained on) is used for nearest neighbors. The features are standardized and cosine similarity is used for nearest neighbor computation. 



\section{Conclusions}

We introduced the self-supervised feature learning task of discriminating natural images from images transformed through local inpainting (LCI), image warping and rotations, based on the principle that trained features generalize better when their task requires detecting global natural image statistics. This principle is supported by substantial experimental evaluation: Trained features achieve SotA performance on several transfer learning benchmarks (Pascal VOC, STL-10, CelebA, and ImageNet) and even slightly outperform supervised training on Places.

\noindent \textbf{Acknowledgements.} This work was supported by the Swiss National Science Foundation (SNSF) grant number 200021\_169622 and an Adobe award.


{\small
\bibliographystyle{ieee_fullname}
\bibliography{refs}
}

\clearpage

\begin{center}
  \textbf{\large Supplementary Material for: \\
Steering Self-Supervised Feature Learning Beyond Local Pixel Statistics}\\
\end{center}

\noindent \textbf{Implementation Details for Limited Context Inpainting.}
We provide additional details regarding the implementation of our Limited Context Inpainting (LCI). The network architecture of the inpainter network $F$ is depicted in Table \ref{tab:inpainter_net}. We used a standard autoencoder architecture with leaky-ReLU activations \cite{maas2013rectifier} and batch normalization \cite{ioffe2015batch}. The architecture of the patch discriminator $D$ is shown in Table \ref{tab:disc_net}. We use spectral normalization \cite{miyato2018spectral} in all the layers of the discriminator. We feed a pair of real or generated patches as input to the discriminator by concatenating them along the channel dimension. We found this to result in more diverse patch inpaintings and more stable training. This technique was also proposed by \cite{lin2018pacgan}. 

\begin{table}[h]
\centering
\caption{Architecture of the inpainter network $F$ used for LCI. Layers in parenthesis are included on ImageNet and excluded for the experiments on STL-10 and CelebA. }
\label{tab:inpainter_net}
\begin{tabular}{@{}c@{}}
\toprule
\textbf{Inpainter Network $F$}                     \\ \midrule
conv $3\times3$ stride=1 \texttt{leaky-ReLU} 48     \\
conv $4\times4$ stride=2 BN \texttt{leaky-ReLU} 96  \\
conv $4\times4$ stride=2 BN \texttt{leaky-ReLU} 192 \\
( conv $4\times4$ stride=2 BN \texttt{leaky-ReLU} 384 ) \\
( deconv $4\times4$ stride=2 BN \texttt{leaky-ReLU} 192 )          \\
deconv $4\times4$ stride=2 BN \texttt{leaky-ReLU} 96          \\
deconv $4\times4$ stride=2 BN \texttt{leaky-ReLU} 48          \\
deconv $3\times3$ stride=1 \texttt{tanh} 3                 \\ \bottomrule
\end{tabular}
\end{table}
\begin{table}[h]
\centering
\caption{Architecture of the patch discriminator network $D$ used for LCI. Layers in parenthesis are included on ImageNet and excluded on STL-10 and CelebA.  }
\label{tab:disc_net}
\begin{tabular}{@{}c@{}}
\toprule
\textbf{Patch Discriminator $D$}       \\ \midrule
conv $3\times3$ stride=1 SN \texttt{leaky-ReLU} 64     \\
conv $4\times4$ stride=2 SN \texttt{leaky-ReLU} 64  \\
conv $3\times3$ stride=1 SN \texttt{leaky-ReLU} 128     \\
conv $4\times4$ stride=2 SN \texttt{leaky-ReLU} 128 \\
conv $3\times3$ stride=1 SN \texttt{leaky-ReLU} 256     \\
( conv $4\times4$ stride=2 SN \texttt{leaky-ReLU} 256 ) \\
( conv $3\times3$ stride=1 SN \texttt{leaky-ReLU} 512 )     \\
Global 2D Average Pooling \\
fully-connected SN \texttt{linear} 1                \\ \bottomrule
\end{tabular}
\end{table}

\noindent \textbf{ResNet Experiments on STL-10.}
We performed additional experiments with a more modern network architecture on STL-10. We followed the setup of \cite{ji2019invariant} and trained a ResNet-34 \cite{he2016deep} for 200 epochs on the 100K unlabelled training images of STL-10. We then fine-tuned the network for 300 epochs on the 5K labelled training images and evaluate on the 8K test images. The training parameters are the same as in our experiments with AlexNet. We used data augmentation and multi-crop evaluation similar to \cite{ji2019invariant}. Results and a comparison to prior work is shown in Table \ref{tab:stl}. 

\begin{table}[t]
\centering
\caption{ Comparison of test-set accuracy on STL-10 with other published results. Note that the methods do not all use the same network architecture.  }
\label{tab:stl}
\begin{tabular}{@{}l@{\hspace{3.em}}c@{}}
\toprule
\textbf{Method}                                          			& \textbf{Accuracy} \\ \midrule
Dosovitskiy \etal  \cite{dosovitskiy2014discriminative}   	&  74.2\% 	\\
Dundar \etal \etal \cite{dundar2015convolutional} 					&  74.1\%	\\ 
Hjelm \etal  \cite{hjelm2018learning} 					&  77.0\% 	\\ 
Huang \etal \cite{Huang_2016_CVPR}  					&  76.8\%  	\\ 
Jenni \& Favaro  \cite{jenni2018self} 					&  80.1\% 	\\ 
Ji \etal  \cite{ji2019invariant} 					&  \underline{ 88.8\% }	\\ 
Oyallon \etal  \cite{oyallon2017scaling} 					&  87.6\% 	\\ 
Swersky \etal  \cite{swersky2013multi}      			&  70.1\%  	\\
Zhao \etal \cite{zhao2015stacked} 					&  74.3\% 	\\ 
\midrule

Ours  &  \textbf{91.8\%} \\
\bottomrule
\end{tabular}
\end{table}

\noindent \textbf{Details of the Evaluation Protocol.}
 For the linear classifier experiments on ImageNet and Places we followed the protocol established by \cite{zhang2016colorful} and train linear classifiers on fixed features extracted at different layers of the network. Feature maps are spatially resized via average-pooling such that they contain approximately 9K units. Training parameters of the linear classifiers are identical to the prior SotA \cite{Feng_2019_CVPR}. Concretely, linear classifiers are trained for 65 epochs using SGD+Momentum with an initial learning rate of 0.1 which we decay to 0.01 after 5 epochs, 0.002 after 25 epochs and 0.0004 after 45 epochs.

\begin{figure}[]
    \centering
    \includegraphics[width=0.95\linewidth]{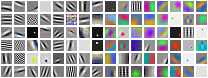}
    \caption{ Features learned in the first conv layer of an AlexNet trained with our method.}
    \label{fig:conv1}
\end{figure}

\begin{figure}[]
    \centering
    \includegraphics[height=1cm,trim={0px 0px 576px 0px},clip]{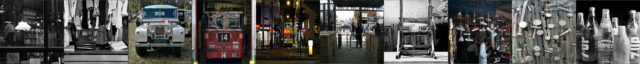}
    \includegraphics[height=1cm,trim={64px 0px 128px 0px},clip]{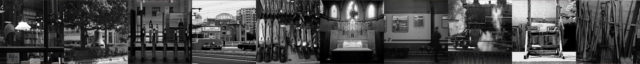}
    \hspace*{1cm} \includegraphics[height=1cm,trim={64px 0px 128px 0px},clip]{figures/nn_sl_7.png}
    \includegraphics[height=1cm,trim={0px 0px 576px 0px},clip]{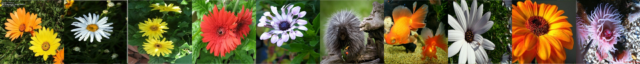}
    \includegraphics[height=1cm,trim={64px 0px 128px 0px},clip]{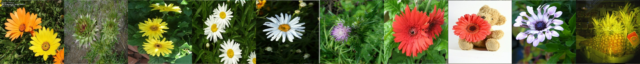}
    \hspace*{1cm} \includegraphics[height=1cm,trim={64px 0px 128px 0px},clip]{figures/nn_sl_11.png}
    \includegraphics[height=1cm,trim={0px 0px 576px 0px},clip]{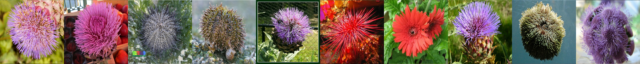}
    \includegraphics[height=1cm,trim={64px 0px 128px 0px},clip]{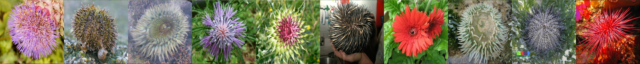}
    \hspace*{1cm} \includegraphics[height=1cm,trim={64px 0px 128px 0px},clip]{figures/nn_sl_3.png}
    
    \includegraphics[height=1cm,trim={0px 0px 576px 0px},clip]{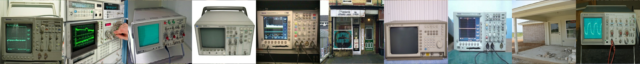}
    \includegraphics[height=1cm,trim={64px 0px 128px 0px},clip]{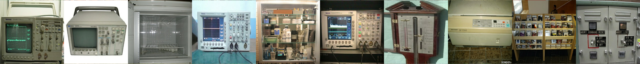}
    \hspace*{1cm} \includegraphics[height=1cm,trim={64px 0px 128px 0px},clip]{figures/nn_sl_13.png}
    \includegraphics[height=1cm,trim={0px 0px 576px 0px},clip]{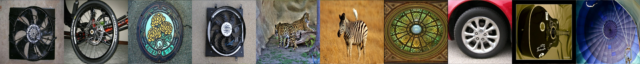}
    \includegraphics[height=1cm,trim={64px 0px 128px 0px},clip]{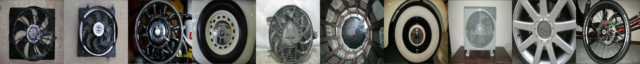}
    \hspace*{1cm} \includegraphics[height=1cm,trim={64px 0px 128px 0px},clip]{figures/nn_sl_14.png}
    \includegraphics[height=1cm,trim={0px 0px 576px 0px},clip]{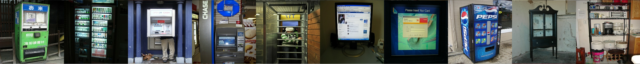}
    \includegraphics[height=1cm,trim={64px 0px 128px 0px},clip]{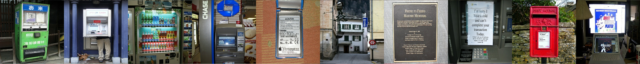}
    \hspace*{1cm} \includegraphics[height=1cm,trim={64px 0px 128px 0px},clip]{figures/nn_sl_15.png}
    \includegraphics[height=1cm,trim={0px 0px 576px 0px},clip]{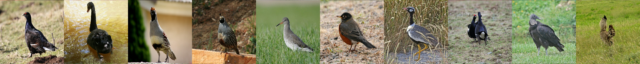}
    \includegraphics[height=1cm,trim={64px 0px 128px 0px},clip]{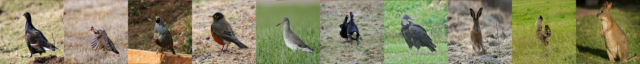}
    \hspace*{1cm} \includegraphics[height=1cm,trim={64px 0px 128px 0px},clip]{figures/nn_sl_16.png}
    \includegraphics[height=1cm,trim={0px 0px 576px 0px},clip]{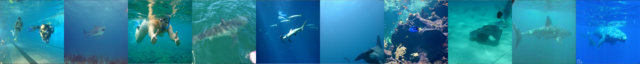}
    \includegraphics[height=1cm,trim={64px 0px 128px 0px},clip]{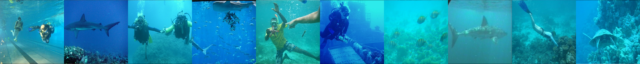}
    \hspace*{1cm} \includegraphics[height=1cm,trim={64px 0px 128px 0px},clip]{figures/nn_sl_17.png}
    \includegraphics[height=1cm,trim={0px 0px 576px 0px},clip]{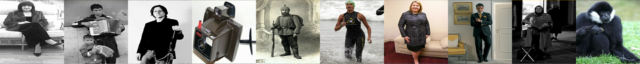}
    \includegraphics[height=1cm,trim={64px 0px 128px 0px},clip]{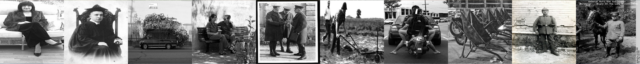}
    \hspace*{1cm} \includegraphics[height=1cm,trim={64px 0px 128px 0px},clip]{figures/nn_sl_18.png}
    \includegraphics[height=1cm,trim={0px 0px 576px 0px},clip]{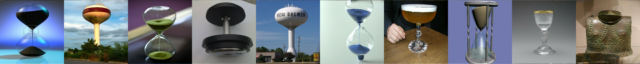}
    \includegraphics[height=1cm,trim={64px 0px 128px 0px},clip]{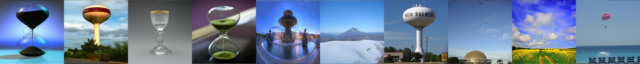}
    \hspace*{1cm} \includegraphics[height=1cm,trim={64px 0px 128px 0px},clip]{figures/nn_sl_19.png}
    \caption{Additional results for nearest neighbor retrieval. The left-most column shows the query image. Odd rows: Retrievals with our features. Even rows: Retrievals with features learned using ImageNet labels.}
    \label{fig:NN_supp}
\end{figure}

\begin{figure}[]
    \centering
    \includegraphics[width=0.24\linewidth]{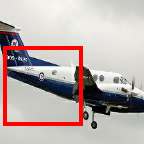}
    \includegraphics[width=0.24\linewidth]{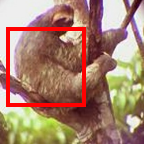}
    \includegraphics[width=0.24\linewidth]{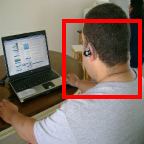}
    \includegraphics[width=0.24\linewidth]{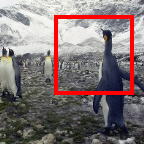}
    \includegraphics[width=0.24\linewidth]{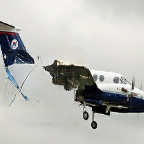}
    \includegraphics[width=0.24\linewidth]{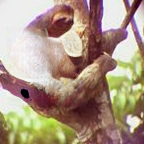}
    \includegraphics[width=0.24\linewidth]{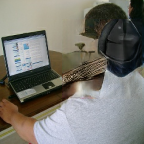}
    \includegraphics[width=0.24\linewidth]{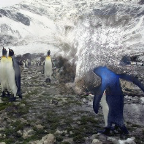} \\
    \vspace{0.08cm}
    \includegraphics[width=0.24\linewidth]{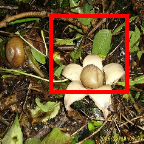}
    \includegraphics[width=0.24\linewidth]{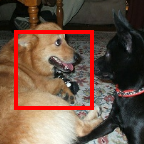}
    \includegraphics[width=0.24\linewidth]{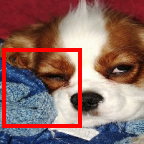}
    \includegraphics[width=0.24\linewidth]{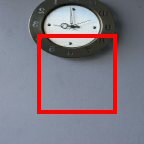}
    \includegraphics[width=0.24\linewidth]{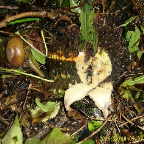}
    \includegraphics[width=0.24\linewidth]{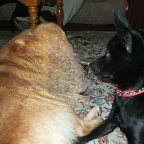}
    \includegraphics[width=0.24\linewidth]{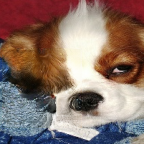}
    \includegraphics[width=0.24\linewidth]{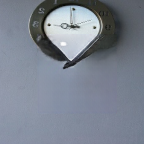} \\
    \vspace{0.08cm}
    \includegraphics[width=0.24\linewidth]{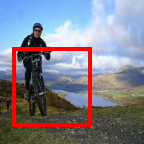}
    \includegraphics[width=0.24\linewidth]{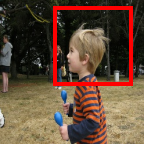}
    \includegraphics[width=0.24\linewidth]{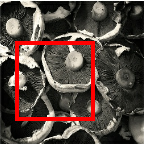}
    \includegraphics[width=0.24\linewidth]{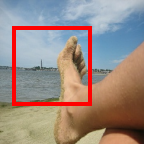}
    \includegraphics[width=0.24\linewidth]{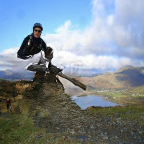}
    \includegraphics[width=0.24\linewidth]{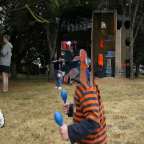}
    \includegraphics[width=0.24\linewidth]{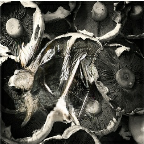}
    \includegraphics[width=0.24\linewidth]{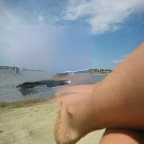} \\
    \vspace{0.08cm}
    \includegraphics[width=0.24\linewidth]{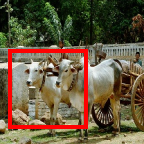}
    \includegraphics[width=0.24\linewidth]{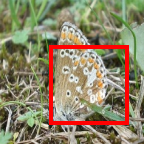}
    \includegraphics[width=0.24\linewidth]{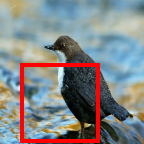}
    \includegraphics[width=0.24\linewidth]{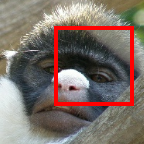}
    \includegraphics[width=0.24\linewidth]{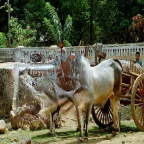}
    \includegraphics[width=0.24\linewidth]{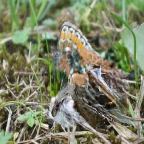}
    \includegraphics[width=0.24\linewidth]{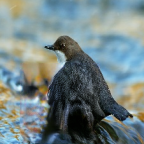}
    \includegraphics[width=0.24\linewidth]{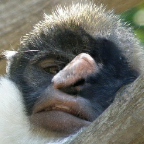} \\
    \vspace{0.08cm}
    \includegraphics[width=0.24\linewidth]{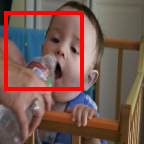}
    \includegraphics[width=0.24\linewidth]{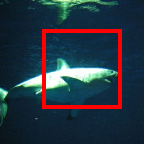}
    \includegraphics[width=0.24\linewidth]{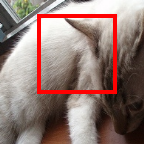}
    \includegraphics[width=0.24\linewidth]{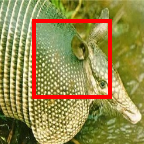}
    \includegraphics[width=0.24\linewidth]{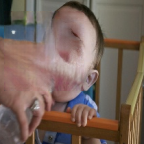}
    \includegraphics[width=0.24\linewidth]{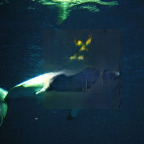}
    \includegraphics[width=0.24\linewidth]{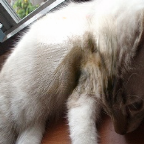}
    \includegraphics[width=0.24\linewidth]{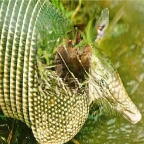} \\
    \caption{We show examples of images transformed with Limited Context Inpainting (LCI). Odd rows: The original training images with the patch used for LCI indicated in red. Even rows: The images after applying LCI. }
    \label{fig:LCI}
\end{figure}

\noindent \textbf{Additional Qualitative Results.}
We visualize the filters learned in the first convolutional layer of an AlexNet after our self-supervised pre-training in Figure \ref{fig:conv1}.
We provide additional results for nearest neighbor retrieval on the ImageNet validation set in Figure \ref{fig:NN_supp}. 
Additionally, we show some examples of LCI transformed images in Figure \ref{fig:LCI}. Note that although the patch-border is in some cases visible, the transformation classifier can not rely on solely detecting these borders, since the examples with autoencoded patches will have similar processing footprints.

\end{document}